%% file: paper_arxiv.tex
\pdfoutput=1
\documentclass[10pt,twocolumn,letterpaper]{article}

\usepackage{cvpr}
\usepackage{times}
\usepackage{epsfig}
\usepackage{graphicx}
\usepackage{amsmath}
\usepackage{amssymb}
\usepackage{multirow}
\usepackage{epigraph}
\usepackage{caption}
\usepackage{color}
\usepackage{subfig}
\usepackage{enumitem}
\usepackage{lipsum}

\usepackage[pagebackref=true,breaklinks=true,letterpaper=true,colorlinks,bookmarks=false]{hyperref}

\newcommand{\lblfig}[1]{\label{fig:#1}}
\newcommand{\e}[1]{{\mathbb E}}

\cvprfinalcopy 

\def\cvprPaperID{} 

\ifcvprfinal\fi
\begin{document}

\title{CIAGAN: Conditional Identity Anonymization Generative Adversarial Networks}

\author{Maxim Maximov{\textbf{\textsuperscript{*}}} \\
    Technical University Munich\\
    \and
    Ismail Elezi{\textbf{\textsuperscript{*}}}\\ 
    University of Venice\\
    \and
    Laura Leal-Taix\'{e}\\
    Technical University Munich\\
}



\twocolumn[{%
\renewcommand\twocolumn[1][]{#1}%
\vspace{-3em}
\maketitle
\thispagestyle{empty}
\vspace{-3em}
\begin{center}
    \centering
    \includegraphics[width=\textwidth, trim={0 0px 0px 0}, clip]{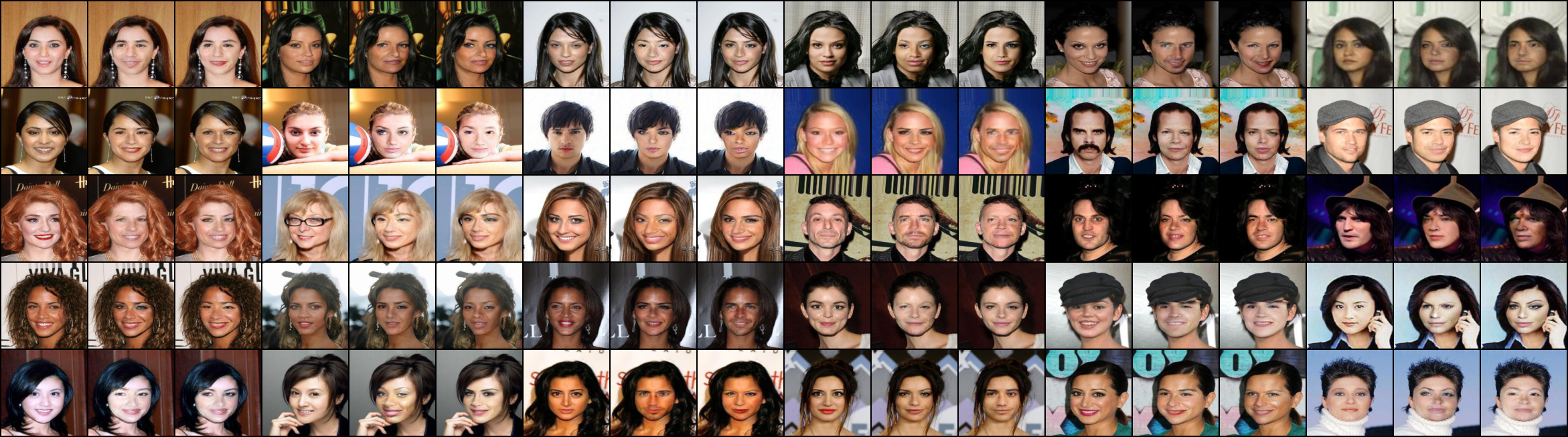}
    \vspace{-1.5em}
    \captionof{figure}{Given a face image, our network anonymizes the face based on the desired identity. In the figure, it can be seen the variability of the generated faces, controlled from a given label. In each triplet, the first image is the real image, while the other two images are different anonymized versions of the real image.}
    \lblfig{teaser}
    \label{teaser}
\end{center}%
}]

\begin{abstract}
The unprecedented increase in the usage of computer vision technology in society goes hand in hand with an increased concern in data privacy. In many real-world scenarios like people tracking or action recognition, it is important to be able to process the data while taking careful consideration in protecting people's identity. 
   We propose and develop CIAGAN, a model for image and video anonymization based on conditional generative adversarial networks. Our model is able to remove the identifying characteristics of faces and bodies while producing high-quality images and videos that can be used for any computer vision task, such as detection or tracking.
   Unlike previous methods, we have full control over the de-identification (anonymization) procedure, ensuring both anonymization as well as diversity. 
   We compare our method to several baselines and achieve state-of-the-art results. To facilitate further research, we make available the code and the models at \url{https://github.com/dvl-tum/ciagan}.
\end{abstract}


\epigraph{“All human beings have three lives: public, private, and \textit{secret}.”}{\textit{Gabriel Garc{\'i}a M{\'a}rquez, \cite{quote}}}

\makeatletter{\renewcommand*{\@makefnmark}{}
\footnotetext{* Authors contributed equally.}\makeatother}

\section{Introduction}

The ubiquitous usage of computer vision technology in society implies the automatic processing of large-scale visual data containing, more often than not, personal data.
While we are eager to take advantage of technology for house monitoring, video conferencing and surveillance, we are not willing to do so by giving away our personal privacy. In fact, data privacy is an increasing concern, and entities such as the European Union have already passed laws such as the General Data Protection Regulations (GDPR)\cite{EUdataregulations2018}, to guarantee data privacy. For computer vision researchers, the creation of high-quality datasets which include people is becoming extremely challenging, as every person in the dataset needs to consent for the usage of his or her image data. Recently, a popular dataset for person re-identification, the Duke MTMC dataset \cite{megapixels}, was taken offline for privacy reasons. 

Our key observation is that many computer vision tasks such as person detection, multiple people tracking, or action recognition, do not need to {\it identify} the people on the videos, they just need to {\it detect} them. Classic anonymization techniques, such as face blurring, significantly alter the image and consequently result in a large detection performance drop.

We propose a model to anonymize (or de-identify) images and videos by removing the {\it identification} characteristics of people, while still keeping necessary features to allow face and body {\it detectors} to work. Importantly, the images should still look realistic to a human observer, but people on them should not be identifiable.
Our proposed approach can be used to anonymize computer vision datasets while retaining necessary information for tasks such as detection, recognition or tracking.
We leverage the generative power of Conditional Generative Adversarial Networks (CGAN) \cite{DBLP:journals/corr/MirzaO14, DBLP:conf/cvpr/IsolaZZE17} to generate anonymized images and videos that look realistic.
In existing GAN-based methods, the image generation process is typically controlled by a random noise vector to generate diverse outputs. Such a random process is not suitable for anonymization purposes, where we need to have guarantees that the identity has actually changed from input to output.
To tackle this issue, we propose a novel identity control discriminator.
Our CIAGAN model fulfills the following important properties that an anonymization system should have: 

\begin{enumerate}[label=(\roman*)]
\item \textbf{Anonymization:} the produced output must hide the identity of the person in the original image. Essentially, we are generating a new fake identity out of the input image.
\item \textbf{Control:} the fake identity of the generated images is governed by a control vector, so we have full control over the real person-fake identity mapping. 
\item \textbf{New identities:} the generated images must contain only new identities not present in the training set.
\item  \textbf{Realistic:} output images must look realistic in order to be used by state-of-the-art detection and recognition systems. 
\item \textbf{Temporal consistency:} temporal consistency and pose preservation in videos should be ensured for tasks like people tracking or action recognition. 
\end{enumerate}

By satisfying the aforementioned five properties, we ensure the anonymization of images and videos and the protection of data privacy. At the same time, our method guarantees that detectors will be able to use the anonymized data, as our experiments demonstrate.

Our {\bf contribution} in this work is four-fold: 
\begin{itemize}
    \item We propose a general framework that is suitable for the anonymization of people in images and video streams.
   
    \item We show that images anonymized by our method can be used by existing detection and recognition systems.
   
    \item We demonstrate state-of-the-art results in several datasets, while at the same time qualitatively demonstrate diversity and control over the generated images.
   
    \item We perform a comprehensive ablation study, showing the importance of each building block of our model.
\end{itemize}

\section{Related Work}

\paragraph{Face Generation}
Since the advent of Generative Adversarial Networks \cite{DBLP:conf/nips/GoodfellowPMXWOCB14, DBLP:journals/corr/RadfordMC15}, generation of realistic faces has been an active research area \cite{DBLP:conf/nips/LiuT16, DBLP:conf/iclr/KarrasALL18, DBLP:journals/corr/abs-1812-04948}. Current state-of-the-art models \cite{DBLP:journals/corr/abs-1812-04948} are able to generate high-resolution face images by progressively training large convolutional neural networks. Diversity in appearance, race, hair and eye color is achieved by adaptive instance normalization \cite{DBLP:conf/iccv/HuangB17}. 
Despite their impressive quality, by conditioning on random noise and having no information about the original face, those methods do not have control over the pose of the generated face. Thus the blending of the face with the other parts of the body is challenging and remains an open research problem. Consequently, their usability in anonymization applications is limited. 

\begin{figure*}
\begin{center}
\includegraphics[width=0.85\textwidth, trim={0 0px 0px 0px}, clip]{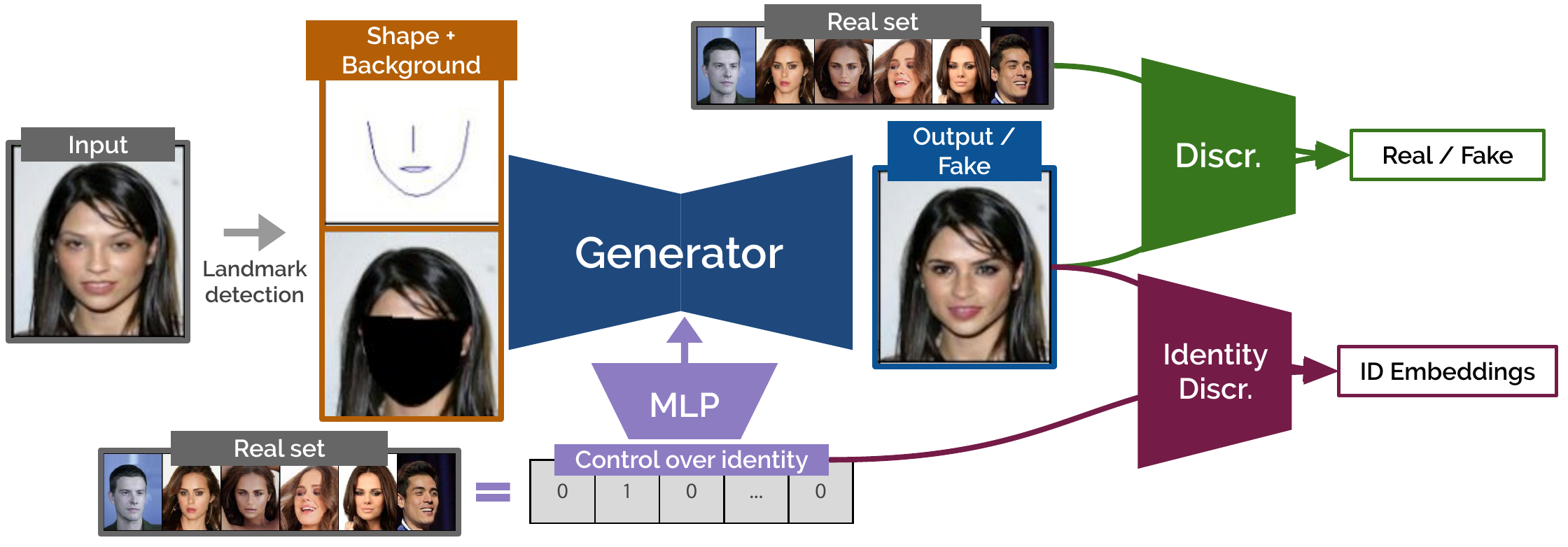}
\end{center}
\vspace{-.5cm}
   \caption{\small Our CIAGAN model takes as input the image, its landmark, the masked face and the desired identity. The generator is an encoder-decoder model where the encoder embeds the image information into a low dimensional space. The identity given as a one-hot label is encoded via a transposed convolutional neural network and is fed in the bottleneck of the generator. Then the decoder decodes the combined information of source images and the identities into a generated image. The generator plays an adversarial game with a discriminator in a standard GAN-setting. Finally, we introduce an identity discriminator network, whose goal is to provide a guiding signal to the generator about the desired identity of the generated face.}
\label{fig:short}
\end{figure*}

\vspace{-0.4cm}
\paragraph{Image-to-Image and Video-to-Video Translation}
Pix2Pix networks \cite{DBLP:conf/cvpr/IsolaZZE17} and their unsupervised variant \cite{DBLP:conf/iccv/ZhuPIE17} have shown impressive results on cross-domain image translation, \eg, from winter to summer. 
Nevertheless, it is not clear how suitable they are on making subtle but important changes in images coming from the same domain, such as faces or bodies. 
Closely related, there has been recent work on ensuring temporal consistency across videos for the task of face translation \cite{DBLP:conf/nips/Wang0ZYTKC18}. To ensure temporal consistency, \cite{DBLP:conf/nips/Wang0ZYTKC18} condition the generator on both the previous real and generated frames, in addition to the estimated optical flow between the frames. While the work demonstrates smooth temporal consistency, generated faces usually resemble closely the original identities, and are not suitable for the anonymization task.

\vspace{-0.4cm}
\paragraph{Face Anonymization}
Until recently, face anonymization has been achieved by pixelization, blurring or masking the faces. Alternatively, \cite{DBLP:conf/aaai/RyooKY18} proposes to use segmentation-based methods. Being based on heuristics rather than being learned, there is no guarantee that these operations are optimal for the task of de-identification. 
Critically, these methods often make faces undetectable and therefore unusable in standard computer vision pipelines.
We advocate the usage of machine \textit{learning} to achieve anonymization that preserves important features that are necessary for computer vision tasks such as detection and tracking. This was previously explored in \cite{DBLP:conf/eccv/RenLR18, DBLP:conf/cvpr/SunMOGSF18, DBLP:journals/corr/abs-1909-04538, DBLP:conf/eccv/SunTXFTS18, DBLP:conf/iccv/Gafni19}.
However, all these works come with important limitations. 
The faces generated by \cite{DBLP:journals/corr/abs-1909-04538} are still, in general, identifiable by humans. 
\cite{DBLP:conf/eccv/RenLR18} exhibits a similar issue and, furthermore, the method has no control over the generation process, with every identity being mapped to the same fake identity. 
The work of \cite{DBLP:conf/cvpr/SunMOGSF18} focuses on altering facial landmarks which can lead to unnatural looking results. Furthermore, their method does not have explicit control over generated appearance. 
The results of \cite{DBLP:conf/eccv/SunTXFTS18} are visually appealing, but due to an optimization procedure for face alignment, the method is not computationally efficient. Additionally, the method is designed to work only with faces since it is based on a parametric face model, so it is not straightforward to extend it to other domains such as full human bodies.

The state-of-the-art method is that of \cite{DBLP:conf/iccv/Gafni19}, where authors show good qualitative results and an unprecedented de-identification rate. 
However, while the generated images can fool recognition systems,
in general, humans can recognize the identity of the presented faces.
More critically, none of these methods except for \cite{DBLP:journals/corr/abs-1909-04538} and \cite{DBLP:conf/iccv/Gafni19} make any attempt at working with videos. 
\cite{DBLP:journals/corr/abs-1909-04538} shows limited experiments in video processing, but the temporal consistency is not well-preserved. 
\cite{DBLP:conf/iccv/Gafni19} shows very good temporal consistency, but as in the case of images, some identities are clearly not anonymized and easy to be spotted by a human eye.
Additionally, they lack control and diversity to show differently anonymized outputs for the same input face.
Our CIAGAN model provides a general framework in anonymizing both images and videos. By providing the labels for the desired generated identities and mixing styles of different identities, we have direct control over the de-identification process. This results in not only high-quality generated images but also with a much higher variability between the images of the same identity (see Fig. \ref{teaser}).

\input{methodology2.tex}

\input{experiments2.tex}

\section{Conclusions and Future Work}
Data privacy in images and videos is a serious problem. As computer vision researchers, we aim to do our part from the technological side. In this paper, we proposed a framework for face and body anonymization in images and videos. Our novel CIAGAN model is based on conditional generative adversarial networks, and faces are anonymized based on a guiding identity signal provided by a siamese network. We have shown that our method outperforms the state-of-the-art in de-identification while showing large diversity in the generated images.

A weakness of all current de-identification methods \cite{ren2018privacy, DBLP:conf/cvpr/SunMOGSF18, DBLP:conf/iccv/Gafni19} is that they need the original faces to be initially detected before they can be anonymized. Consequently, any face that has not been detected can not be anonymized. Therefore, these methods are not deployable in systems where anonymization must be guaranteed. 
Our method suffers from a similar issue as it is dependent on landmark detection. As future work, we plan on working on full image anonymization and further eliminate the need for landmark detection in order to be able to handle extreme poses.

\noindent\textbf{Acknowledgements.} This research was partially funded by the Humboldt Foundation through the Sofja Kovalevskaja Award. We thank Aljosa Osep, Tim Meinhardt and Patrick Dendorfer for their helpful insights.

{\small
\bibliographystyle{ieee_fullname}
\bibliography{egbib}
}

\clearpage
\input{supp_arxiv}

\end{document}

%% file: methodology2.tex
\section{CIAGAN}
In this section, we detail our methodology to anonymize images and videos. Our proposed Conditional Identity Anonymization Generative Adversarial Network (CIAGAN) leverages the power of generative adversarial networks to produce realistic images.
In order to have control over the identity generation process and guarantee anonymization, we propose a novel identity discriminator to train CIAGAN.
For the remainder of the section, we will specifically refer to face anonymization, even though the method is directly applicable to full bodies.

\subsection{Method overview}
We show a full diagram of CIAGAN in Fig. \ref{fig:short}. The main components of our method are the following:

\noindent\textbf{Pose preservation and temporal consistency.} We propose using a landmark-based representation of the input face (or body). This has two advantages: it ensures pose preservation which is especially useful for, \eg, tracking; it provides a simple but efficient way to maintain temporal consistency when working with videos.
   
\noindent\textbf{Conditional GAN.} We exploit the generative power of GANs to produce realistic-looking results. It is important that standard detection and tracking systems can be applied to generated images without accuracy loss. Naturally, the realistic-looking generated faces are easily detectable.
We achieve pose preservation by conditioning on the landmark representation.
We train the conditional GAN in an adversarial game-theoretical way, where the discriminator judges the realism of the images generated by the generator.

\noindent\textbf{Identity guidance discriminator.} 
We propose a novel module that controls the identifying characteristics that the generator injects to create the new image.
The identity discriminator and the generator play a collaboration game where they work together to achieve their common goal of generating realistic anonymized images.
We now provide a more detailed description of the three modules of our method.

\subsection{Pose preservation and temporal consistency}
\label{sec:pose}

Several de-identification methods \cite{DBLP:conf/eccv/RenLR18, DBLP:conf/iccv/Gafni19} take as input the RGB image of the face to be anonymized. It is not surprising that there typically is some leakage of face information to the generated image. Thus, while those methods produce high-quality images, the identity of the generated faces is not fully anonymized and can be often identified by a human. 
\vspace{-0.5cm}
\paragraph{Landmark image.} To make sure our generated faces can not be linked to the original identity, we propose to work with an abstraction of faces. More precisely, we use face landmark images.
This has two advantages: (i) the landmark image contains a sparse representation of the face with little identity information left, avoiding identity leakage, and (ii) the generator is conditioned on the face shape, allowing us to preserve the input pose in the output. 
This is especially important as we intend to use the generated images and videos as inputs to computer vision algorithms. In many vision applications, \eg tracking, methods typically leverage the pose of the face or the body. Thereby, assuring the method will not change the pose of the anonymized face or body is extremely useful.
To hide as much identity information as possible but still preserve the pose, instead of using all $68$ landmarks \cite{dlib09}, we use only the face silhouette, the mouth, and the bridge of the nose (see Fig. \ref{fig:short}). 
This allows the network the freedom of choosing several facial characteristics, such as eye distance or eye shape, while at the same time, preserving expressions that are dependent on the mouth region, e.g. smiling or laughing, and the global pose is given by the nose position.
The landmarks are represented as a binary image which is fed as input to the generator.

\vspace{-0.5cm}
\paragraph{Masked background image.} Our goal is to generate only the face region of an image and inpainting it to the original image background. This allows our algorithm to focus the learning power to the generation of faces (and not that of background), and at the same time guarantees that we do not have background changes that could interfere with the detection or tracking algorithms. To achieve this, we provide the generative model with the {\it masked background image} together with the {\it landmark image}.
The masked background image still contains the forehead region of the head. Once the generator has access to this information, it can learn to match the skin appearance of the generated face to the forehead skin color. This leads to overall better quality of visual results. 
In cases there are multiple faces in the same image, we detect each face on the image and sequentially apply our anonymization framework. 

Our pipeline can be additionally used for full-body anonymization by simply replacing the mask image with a segmentation mask representing the silhouette of the body. In our case, we do not use body joints as a replacement for the landmark image, as the silhouette of a person suffices as a pose prior.

\noindent{\bf Temporal consistency.} 
In order to work on videos, any de-anonymization pipeline must ensure the temporal consistency of generated images on the video sequence. The state-of-the-art video translation model \cite{DBLP:conf/nips/Wang0ZYTKC18} ensures temporal consistency by using a discriminator that is conditioned on the optical flow between corresponding frames. The optical flow is computed via an external neural network \cite{DBLP:conf/iccv/DosovitskiyFIHH15}, which makes the framework both complex and computationally expensive. 
In our work, due to the nature of our input representation, we obtain temporal consistency for free. 
The landmarks of every frame are smoothed using a spline interpolation over neighboring frames. Therefore, we provide the same framework for both images and videos, with the only difference being the computationally cheap interpolation done at inference time.

\subsection{Conditional generative adversarial networks}
\noindent{\bf GAN framework.} 
In simple terms, GANs combine two neural networks: a generator $G$ whose goal is to generate realistic-looking samples, and a discriminator $D$ whose goal is to differentiate between the real samples and the generated ones. The networks are trained in an adversarial manner with $D$ being trained to maximize the probability of assigning the correct label to both training and generated examples, and $G$ being trained to minimize the probability of $D$ predicting the correct labels for the generated samples. In other words, $D$ learns to separate the real samples from the generated ones, while $G$ learns to fool $D$ into classifying the generated samples as real ones.
It is well-known that GAN training is hard and requires many tricks \cite{DBLP:conf/iccv/MaoLXLWS17, DBLP:conf/nips/GulrajaniAADC17, DBLP:conf/nips/ChenCDHSSA16}. In this work, we choose to train CIAGAN with the LSGAN loss function \cite{DBLP:conf/iccv/MaoLXLWS17}.
The idea of using least-squares loss function for GAN training is simple yet powerful: the least-squares loss function is able to move the fake samples toward the decision boundary, as it penalizes also samples that are correctly classified but still lie far away from real samples. This is opposed to the cross-entropy loss that mostly penalizes wrongly classified samples. Based on this property, LSGANs are able to generate samples that are closer to real data.

The objective function of the discriminator in the LSGAN setting is defined as follows:

\vspace{-0.3cm}
\begin{equation}
\begin{split}
\min_D V_{LSGAN} (D) = \frac{1}{2} \e_{x \sim p_{data}(x)} [(D(x) - b)^2] + \\
\frac{1}{2} \e_{z \sim p_{z}(z)} [(D(G(z)) - a)^2],
\end{split}
\end{equation}

where $a$ and $b$ are the labels for the fake and the real data.

\vspace{0.3cm}

The loss of the generator is defined as:
\begin{equation}
\min_G V_{LSGAN} (G) = 
\frac{1}{2} \e_{z \sim p_{{z}} ({z})} [(D(G(z)) - b)^2],
\end{equation}

Without loss of generality, LSGAN can be replaced with any of the other common loss functions used for GAN training \cite{DBLP:conf/nips/GulrajaniAADC17, DBLP:conf/nips/ChenCDHSSA16}.

\noindent{\bf Conditional GAN.} In the classic GAN training setting, a vector of random noise is given as input to the generator in order to provide variability to the generated images. %
In our case, it is necessary for the generated faces to be aligned with the landmarks from the input image, for pose preservation and temporal consistency. Furthermore, we also need to blend the generated faces seamlessly with a background. To this end, we use the conditional GAN framework \cite{DBLP:conf/cvpr/IsolaZZE17}, where we condition the generator with a landmark and a masked image (background), as explained in Section \ref{sec:pose}. 
The generator uses an encoder-decoder architecture \cite{DBLP:conf/cvpr/LongSD15}. The encoder brings the landmark and the masked image to a low-dimensional representation (bottleneck), where it is combined with identity representation while the decoder takes combined representation and upsamples it to generate the anonymized RGB image.

\label{guidance}
\subsection{Identity guidance} 
With the two modules explained above, our model learns to generate faces that look realistic and preserve the pose of original images. However, if the entire variability in image generation is provided by the landmark input, the network quickly overfits on the training set, effectively doing only image reconstruction. 
By doing so, it generates faces that are very similar to those in the training dataset, forfeiting the final anonymization goal. 
To solve this problem, we introduce a novel identity guidance discriminator.
More precisely, for every given real image, we randomly choose the desired identity of its corresponding generated image. This identity -- represented as on one hot-vector -- is given as input to a transposed convolutional neural network. The network produces a parametrized version of the identity and feeds it into the bottleneck of the generator. In this way, the generator learns to generate a face with some of the characteristics of the desired identity while keeping the pose of the real image. In other words, the generated image is a composition of both the landmark identity and the desired identity. The identity of the generated image must not be the same as any of the real identities in order for the generated image to be not recognizable. 

The identity discriminator is designed as a siamese neural network pretrained using Proxy-NCA loss \cite{DBLP:conf/iccv/Movshovitz-Attias17}. Pre-training is done using real images, where the discriminator is trained to bring together features coming from the images that belong to the same identity.
During the GAN training, we finetune the siamese network using contrastive loss \cite{bromley1994signature}. During this fine-tuning step, we allow the siamese network to bring together the ID representation of the fake images and the real images. 
The identity discriminator and the generator are trained jointly in a collaborative manner. The identity discriminator's goal is to provide a guidance signal to the generator to guide it towards creating images whose representational features are similar to those of a particular identity. 

\vspace{-0.3cm}
\paragraph{The case for multiple object tracking.} 
Of particular interest is the {\it control} on the fake identity generation. We need to be able to keep the same real person-fake identity mapping within a sequence taken from a camera \eg, multiple object tracking, but at the same time change the mapping for different cameras to avoid long-term tracking and potential misuse of the data. To do so, when a person moves from one camera to another, we replace its {\it control} vector with a new one, in turn giving the person a new identity. This is a simple yet powerful way of doing multiple object tracking within the frames taken from a camera, without the undesirable consequence of doing long-term tracking.

%% file: experiments2.tex
\section{Experiments}
In this section, we compare CIAGAN with several classic and learning-based methods commonly used for identity anonymization. Our method achieves state-of-the-art qualitative and quantitative results in different image and video datasets. We also present a set of comprehensive ablation studies to demonstrate the effect of our design choices. 
We first introduce the datasets, evaluation metrics and baselines used throughout this section.\\

\noindent{\bf Datasets.} We perform experiments on $3$ public datasets:
\begin{itemize}
\setlength\itemsep{-0.5em}
\item \noindent{\bf CelebA \cite{liu2015faceattributes}} The dataset consists of $202,599$ face images of $10,177$ unique identities. We use the aligned version where each image is centered on a point in-between person's eyes, and then padded and resized to have $178 \times 218$ resolution, while maintaining original face proportions. Each identity contains up to $35$ photos. We use HOG \cite{DBLP:conf/cvpr/DalalT05} to construct face landmarks for each face. 
\item  \noindent{\bf MOTS \cite{DBLP:conf/cvpr/VoigtlaenderKOL19}} Our method can also be adapted to work in other domains such as full-body anonymization. Instead of using face landmarks we use body segmentation masks. The dataset consist of $3,425$ videos of $1,595$ different people. 
\item  \noindent{\bf Labeled Faces in the Wild (LFW) \cite{LFWTech}} The dataset consists of $6,000$ pair images, split in $10$ different splits, where half of the pairs contains images of same identity, and the remaining pairs consist of images that have different identities.
\end{itemize}

\noindent{\bf Baselines.} We compare to standard anonymization methods as well as learning-based methods.
\begin{itemize}
\setlength\itemsep{-0.5em}
\item \noindent{\bf Simple Anonymization methods.} We use pixelization, blur and masking of faces and compare them with our method.
\item \noindent{\bf Image Translation methods.} We use the popular pix2pix \cite{DBLP:conf/cvpr/IsolaZZE17} and CycleGAN \cite{DBLP:conf/iccv/ZhuPIE17} methods. We use the official code given by the authors, and we present the results in the supplementary material.
\item \noindent{\bf Face Replacement methods.} We compare the results in de-identification with the state-of-the-art results given by \cite{DBLP:conf/iccv/Gafni19}. 
\end{itemize}

\subsection{Implementation details}
We generate the landmarks and the masks using Dlib-ml library \cite{dlib09}. We train our network on $128 \times 128$ resolution images, and use an encoder-decoder U-Net \cite{DBLP:conf/miccai/RonnebergerFB15} architecture for the generator. The identity vector is parametrized by a transposed convolutional neural network containing a fully connected layer followed by multiple transposed convolutional layers. The features coming from the landmark and the identity branches are concatenated in the generator's bottleneck. For the discriminator, we use a standard convolutional neural network that has the same architecture as the identity guidance network. We train our model for $60$ epochs using the ADAM optimizer \cite{DBLP:conf/iclr/Kingma14} with $1e-5$ learning rate. We set the beta hyperparameters $\beta_1$, $\beta_2$ to $0.5$ and $0.9$. The total training time for the model is one day in a single GPU. In the supplementary material, we give the remaining implementation details and the network architectures.

\subsection{Evaluation metrics}

We evaluate all models in face detection and re-identification metrics. 
We perform detection using HOG \cite{DBLP:conf/cvpr/DalalT05} and SSH Detector \cite{DBLP:conf/iccv/NajibiSCD17}. To evaluate the performance of the detectors, we use the percentage of detected faces.
 For re-identification, we train a siamese neural network using Proxy-NCA \cite{DBLP:conf/iccv/Movshovitz-Attias17}. Additionally, we use a pre-trained FaceNet model \cite{DBLP:conf/cvpr/SchroffKP15} based on Inception-Resnet backbone \cite{DBLP:conf/aaai/SzegedyIVA17}. We use the standard Recall@1 evaluation metric for re-identification. It measures the ratio of samples whose nearest neighbor is from the same class. The metric can take values from $0$ to $100$ with $0$ showing perfect de-identification rate and $100$ showing perfect identification rate. Note that in a balanced dataset, a random classifier will produce (on average) a Recall@1 of $1/|C|$ where $C$ is the number of classes.
Finally, we evaluate the visual quality of the generated images quantitatively using the Fr\'echet Inception Distance (FID) \cite{DBLP:conf/nips/HeuselRUNH17}, a metric that compares the statistics of generated samples to those of real samples. The lower the FID, the better, corresponding to more similar real and generated samples. 

\subsection{Ablation Study}

In this section, we perform an ablation study of our method in order to demonstrate the value of our design choices. 
In Table \ref{tab:ablation}, we show several variants of our model. {\it Siamese} indicates our full model, with a siamese identity discriminator, and using landmarks as input. 
{\it Classification} indicates replacing the siamese identity discriminator by a classification network. As we can see, the results of the detection drop more than $35$ percentage points ($pp$). We also show what happens if instead of landmarks, entire face images are provided as input. In these cases, the detection rate drops $1.6pp$, and the FID score increases, showing that the faces are both more difficult to be detected, and have lower visual quality.

\input{tables/table_ablation.tex}
\input{tables/table_detection_id}

\subsection{Quantitative results}
\paragraph{Detection and Recognition.} The first experiment evaluates two important capabilities that an anonymization method should have: high detection rate and low identification rate. That is, we do not want a trained system to be able to find the identity of the newly generated face, but at the same time, we still want a face detector to have a high detection rate.

In Table \ref{tab:detection_id}, we show the detection and identification results of our method compared to the other methods on the CelebA dataset \cite{liu2015faceattributes}. The detection rate of classical HOG \cite{DBLP:conf/cvpr/DalalT05} and deep learning-based SSH \cite{DBLP:conf/iccv/NajibiSCD17} detectors in our anonymized images is at almost 100\%. Blurring methods have a much lower detection rate in the images, while faces in the pixelized images are not detectable at all.

The identification performance drops from more than $70\%$ recall on the original dataset to $1-1.5\%$ recall on our anonymized images. The CIAGAN generated images are almost unrecognizable by any of the two identification systems. Note, pixelization methods reach $0.3\%$ recall, equivalent to random guessing, but that comes at the cost of deleting all content from the image, making both detection and identification impossible. In a setup where we want to further use computer vision algorithms on anonymized data, neither pixelization nor blurring is an option.

\vspace{0.3cm}
\noindent{\bf Recognition based on landmarks.} 
Given that the input to our generator is a landmark image, and not the actual image we want to de-anonymize, one might argue that an identification method that focuses on image pixels is easily fooled by our method, as we saw in Table \ref{tab:detection_id}. 
What happens if CIAGAN is attacked by an identification method trained only on landmarks? Will it still maintain its anonymization capabilities? 
We perform this experiment by training a similar identification method \cite{DBLP:conf/iccv/Movshovitz-Attias17} as before, but with landmarks as the only input. 
We evaluate that by only using landmarks, we can identify up to $30.5\%$ compared to $70.7\%$ when using full images.
However, when the same identifier is evaluated on the landmarks extracted from our anonymized faces, the performance drops to $1.9\%$ recall. 
Even if the original landmarks are used as input to our generation model, CIAGAN uses them only as prior information to fuse it with the information coming from the embedder network. 
%

\input{tables/table_lwf}
%
%
\vspace{0.3cm}

\begin{figure}[t]
  \centering
    \includegraphics[width=0.4\textwidth, trim={0px 0px 0px 0px}, clip]{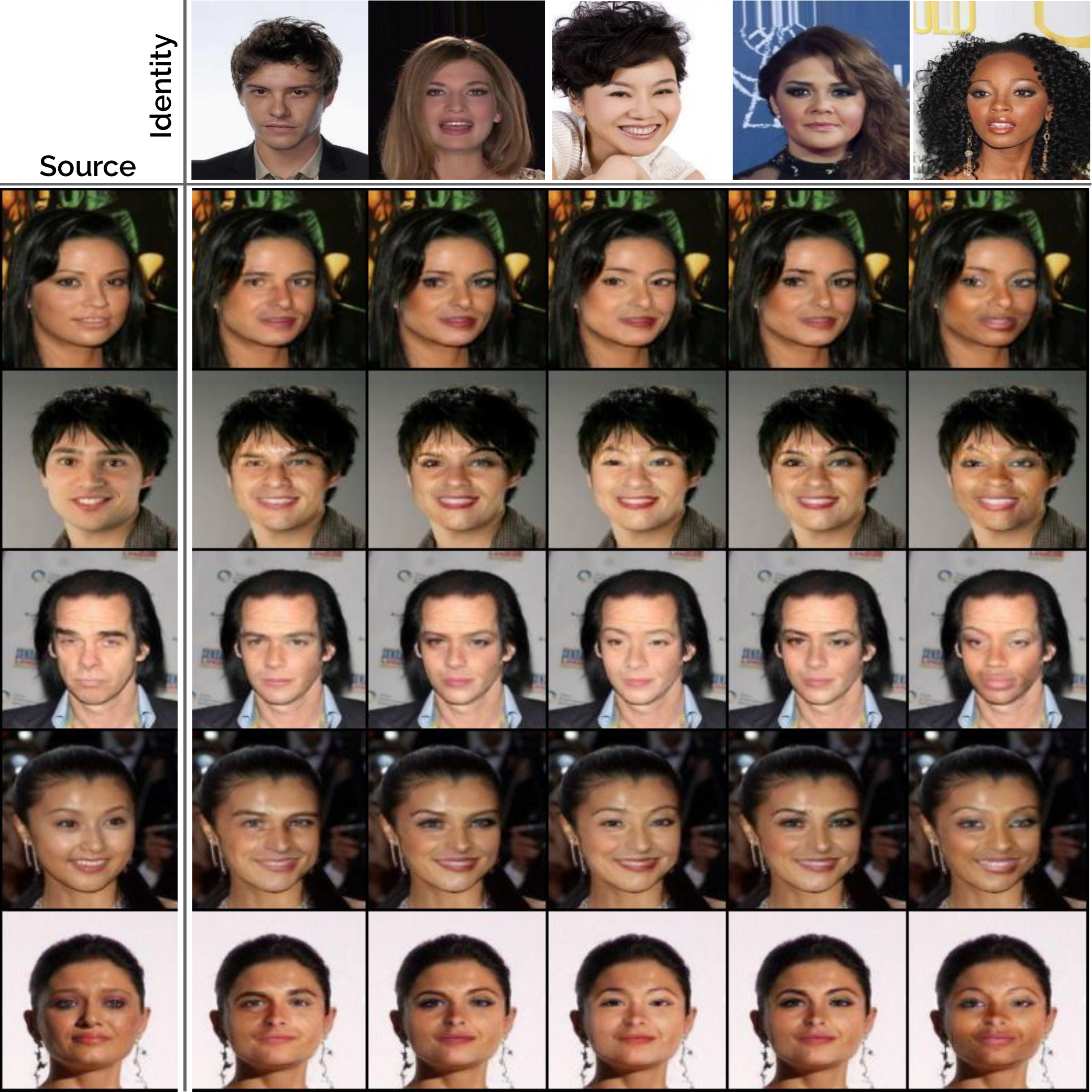}
  \caption{\small Generated faces of our model, where a source image is anonymized based on different identities.}
  \label{fig:combine}
\end{figure}

\noindent{\bf Are we just doing face swapping?} 
In Section \ref{guidance}, we introduced a novel identity guidance network that guides the generator to produce images with similar features to those of a given identity. 
One might argue that by doing so, the generator is learning to only do face swapping, replacing the face of the chosen identity into the landmarks of the source image. 
We show that this is not the case, by evaluating the identification rate of our generated images on the training set of the real images. 
We set the label of the generated images to the labels of their desired identity. If the generator is learning to do only face swapping, then a recognizer would be able to identify correctly all generated images. 
However, we show that this is not the case. Neither FaceNet \cite{DBLP:conf/cvpr/SchroffKP15} nor our model trained in P-NCA \cite{DBLP:conf/iccv/Movshovitz-Attias17} are able to achieve a higher recognition rate than a random guesser. 
Additionally, in Fig. \ref{fig:combine} we present a qualitative experiment, where the first image of each row contains the source images while the first image in each column is a randomly chosen image from the desired identities. The other images are generated. We see that the generated images take high-level characteristics of their given identities (such as race or sex), but differ greatly from the real images of those identities.

\vspace{-0.5cm}
\subsubsection{Comparison to SOTA in de-identification} 
In this section, we compare the de-identification (anonymization) power of our model compared to state-of-the-art \cite{DBLP:conf/iccv/Gafni19}.
We follow their evaluation protocol on the LFW dataset \cite{LFWTech}. The dataset consists of $10$ different splits, where each contains $600$ pairs. A pair is defined as \textit{positive} if both elements share the same identity, otherwise as \textit{negative}. In every split, the first $300$ pairs are positive, and the remaining $300$ pairs are negative. As in \cite{DBLP:conf/iccv/Gafni19}, we anonymize the second image in every pair. 

We use FaceNet \cite{DBLP:conf/cvpr/SchroffKP15} identification model, pre-trained on two public datasets: VGGFace2 \cite{DBLP:conf/fgr/CaoSXPZ18} and CASIA-Webface \cite{DBLP:journals/corr/YiLLL14a}. 
The main evaluation metric is the true acceptance rate: the ratio of true positives for a maximum $0.001$ ratio of false positives.
We present the results in Table \ref{tab:lwf}. The network evaluated in real faces reaches a score of almost $0.99$, nearly perfect identification.
\cite{DBLP:conf/iccv/Gafni19} achieves an impressive anonymization performance by obtaining a score of less than $0.04$ using the networks trained in both datasets. 
%
CIAGAN improves on this result and lowers the identification rate to $0.034$ using the network trained in \cite{DBLP:conf/fgr/CaoSXPZ18} and $0.019$ using the network trained in \cite{DBLP:journals/corr/YiLLL14a}, thus improving anonymization. On average, CIAGAN shows a $10.5\%$ better de-identification rate on the first dataset, and a $45.7\%$ better de-identification rate on the second dataset, while keeping a high detection rate of $99.13\%$. The average performance of $2.65\%$ true positive rate shows that even a near-flawless system would completely fail to find the true identities in our CIAGAN-processed data, showing the strength of our method in achieving image anonymization.

\subsection{Visual quality of the results}

As can be seen in Table \ref{tab:ablation}, our method reaches an FID score of $2.08$.
Simple baselines such as blurring and image translation methods reach a significantly higher (worse) FID score. 
The FID score comparison and qualitative results of the baselines can be found in the supplementary material.

We show a series of qualitative results.
In Fig.~\ref{teaser}, we show the diversity of the generated images, when the control vector of the identity discriminator changes. We see that the generated images take high-level characteristics of the desired identity (such as eye shape, race or sex) while generating realistically looking images.

\begin{figure}[t!]
  \centering
    \includegraphics[width=0.4\textwidth]{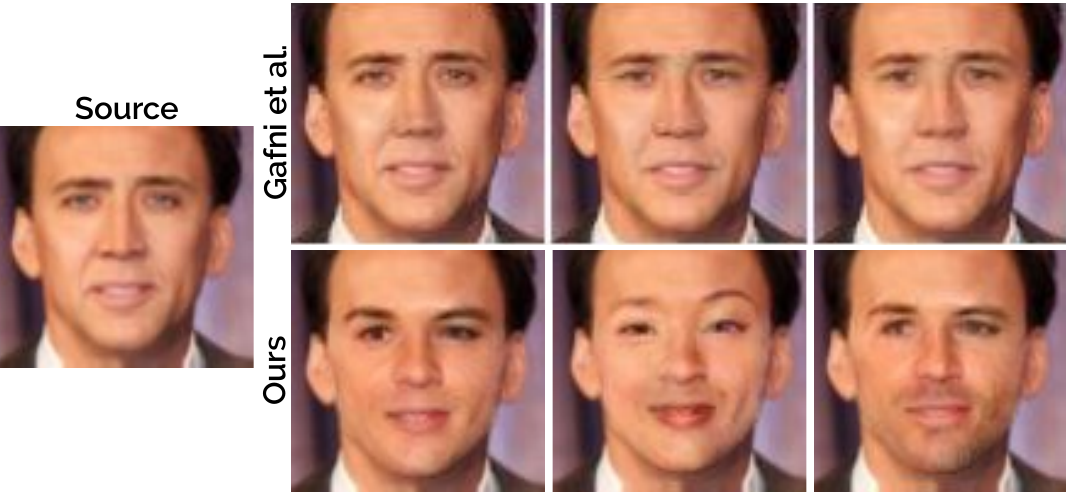}
    \vspace{-0.5em}
  \caption{\small Qualitative comparison with \cite{DBLP:conf/iccv/Gafni19}. The image in the first column is the source image. In the first row, we show the generated images from the framework of \cite{DBLP:conf/iccv/Gafni19}, while in the second image, we show the generated images from CIAGAN.}
  \label{fig:cage}
\end{figure}

In Fig. \ref{fig:cage}, we qualitatively compare our results with those of \cite{DBLP:conf/iccv/Gafni19}. We see that our method not only provides images which are more dissimilar to the source image, but by changing the control vector, our network is able to provide much more diverse images than those of \cite{DBLP:conf/iccv/Gafni19} where the authors gradually change their control parameter $\lambda$ (we can still recognize N. Cage).

\begin{figure}[t!]
  \centering
    \includegraphics[width=0.4\textwidth, trim={0 0px 0px 0}, clip]{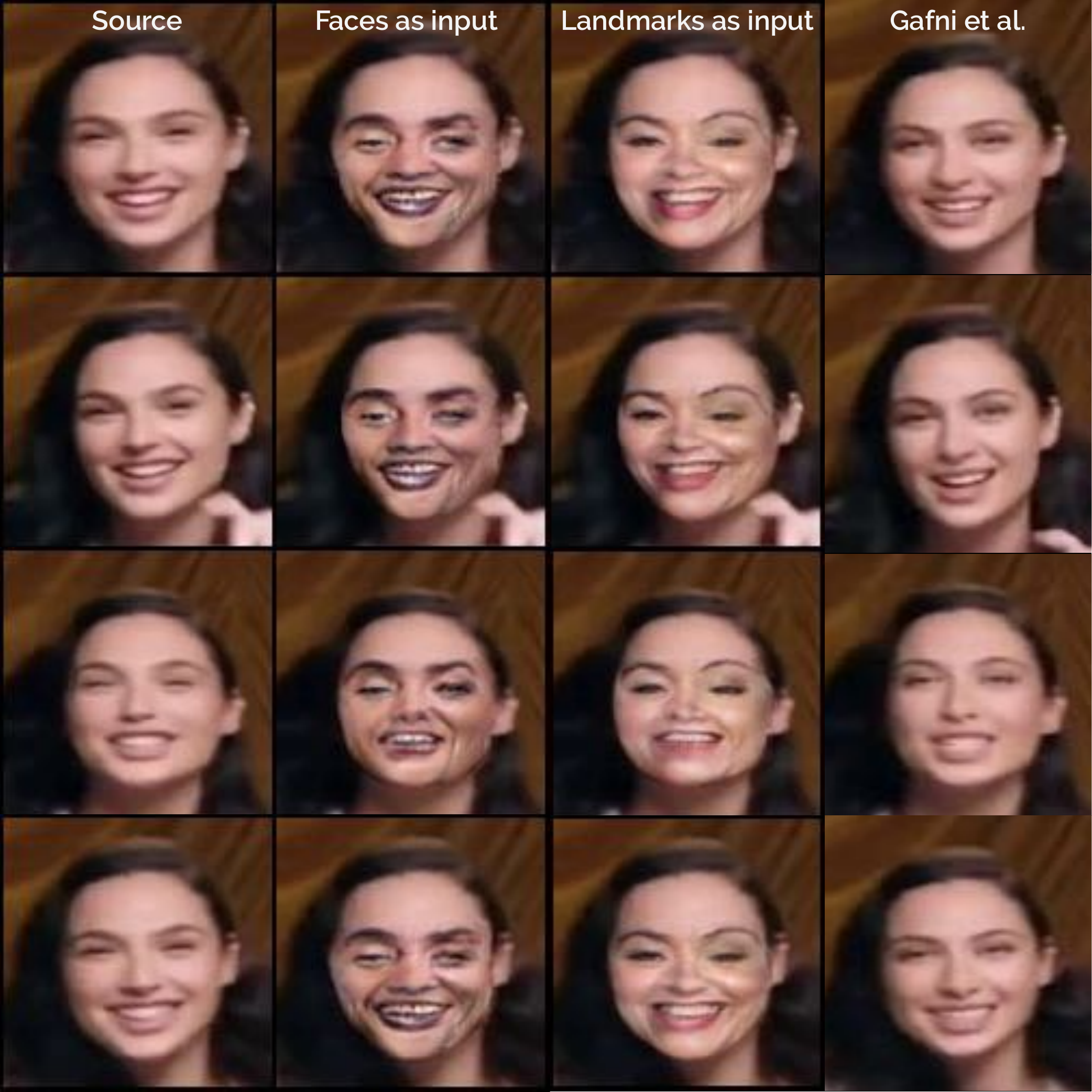}
    \vspace{-0.5em}
  \caption{\small Qualitative comparison with \cite{DBLP:conf/iccv/Gafni19} on temporal consistency. From left to right: original frames; faces generated by our model using faces as input; faces generated by our model using landmarks as input, faces generated by \cite{DBLP:conf/iccv/Gafni19}.}
  \label{fig:temporal_consistency}
\end{figure}

In Fig. \ref{fig:temporal_consistency}, we demonstrate the temporal consistency of our method and compare the results with those of \cite{DBLP:conf/iccv/Gafni19}. We see that the pose is preserved in all cases, yielding excellent temporal consistency. At the same time, we see that the CIAGAN version which works with landmarks produces better looking images than the version trained on full faces.

\begin{figure}[t!]
  \centering
    \includegraphics[width=0.4\textwidth]{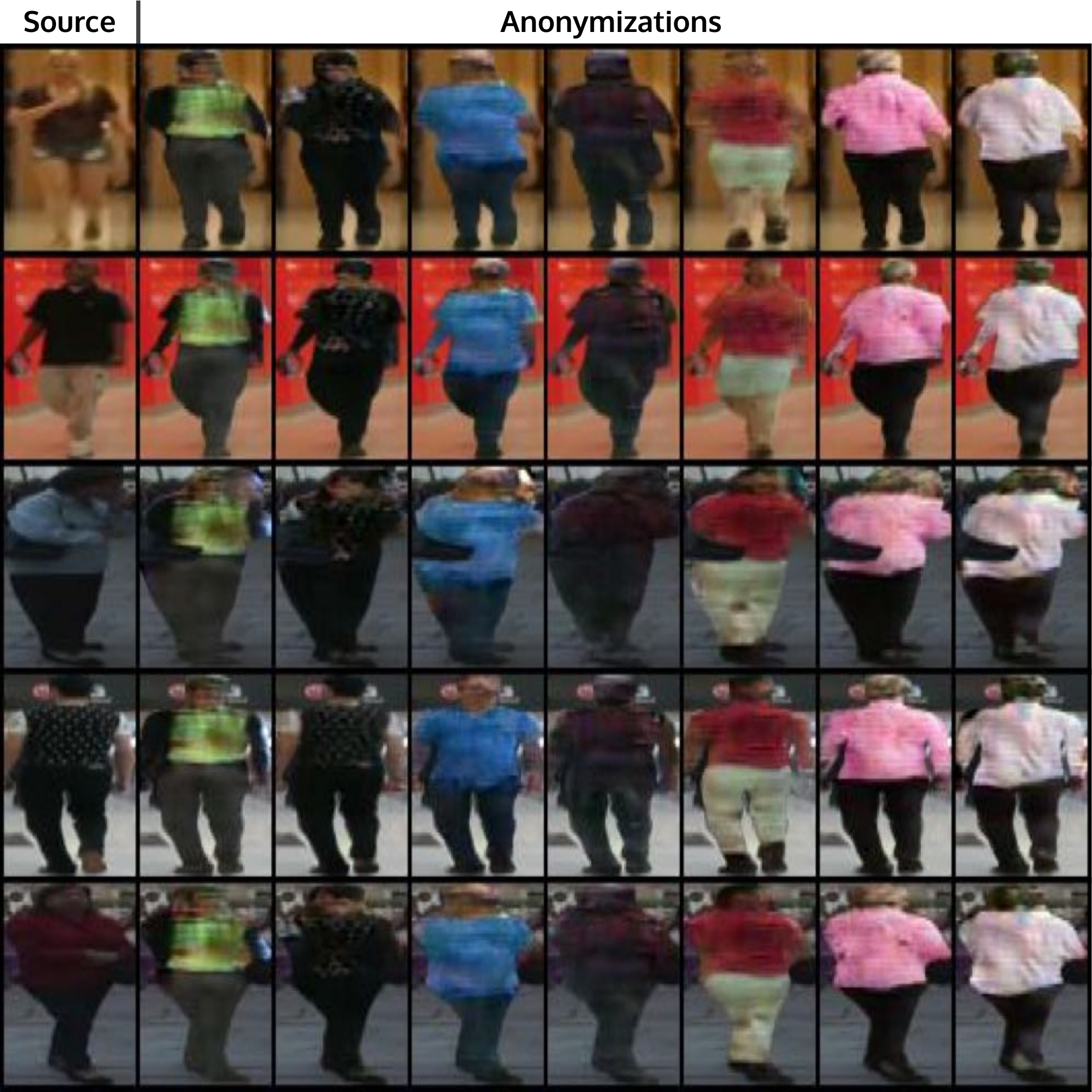}
    \vspace{-0.5em}
  \caption{\small The full-body anonymization results of our framework in MOTS dataset.}
  \label{fig:mots}
\end{figure}

Finally, in Fig. \ref{fig:mots}, we show an experiment on full-body anonymization. The first image in each row is the source image, while the other images are the generated anonymized images. In each case, we see that the generated images keep the same posture as the source image counterpart, but the clothes, colors and other parts of the body are changed. To the best of our knowledge, this is the first time that the same framework has been successfully used to perform both face and body de-identification. 

%% file: tables/table_ablation.tex
\begin{table}[]
\begin{center}
\begin{tabular}{llll}
\hline
{Models} & {Detection ($\uparrow$)} & {Recall@1 ($\downarrow$)} & {FID ($\downarrow$)}  \\ \hline
Siamese               &   \textbf{99.9}    &  1.3  &  \textbf{2.1} \\ 
Classification           &   64.6    &  \textbf{0.4}  &  63.2  \\  
Faces                    &   98.3    &  1.1  &  6.5   \\ \hline

\end{tabular}
\end{center}
\vspace{-0.5cm}
\caption{\small Ablation study of our model. First row presents the result of our model, second row shows the result of the model where the siamese identity guidance network is replaced with a classification network, while the third row shows the result of the model where the generator accepts full face images instead of landmarks.}
\label{tab:ablation}
\end{table}

%% file: tables/table_detection_id.tex
\begin{table}[]
\begin{center}
\begin{tabular}{lllll}
\hline
\multicolumn{1}{c}{\multirow{2}{*}{Models}} & \multicolumn{2}{c}{Detection ($\uparrow$)} & \multicolumn{2}{c}{Identification ($\downarrow$)}   \\ 
\multicolumn{1}{c}{}                        & \multicolumn{1}{c}{Dlib} & \multicolumn{1}{c}{SSH} & \multicolumn{1}{c}{PNCA} & \multicolumn{1}{c}{FaceNet}\\ \hline
Original                 &    100            &   100                 &      70.7        &   65.1          \\
Pixelization 16 by 16                &    0.0            &   0.0          &       \textbf{0.3}        &    \textbf{0.3}          \\ 
Pixelization 8 by 8                  &    0.0            &   0.0          &       0.4        &    \textbf{0.3}          \\ 
Blur 9 by 9                            &    90.6          &   38.6            &      16.9        &   57.2          \\ 
Blur 17 by 17                           &    68.4          &  0.3              &      1.9        &   0.5          \\ \hline
Ours                      &       \textbf{97.8}\footnotemark       &    \textbf{97.4}\footnotemark[\value{footnote}]                 &       1.3        &    1.0        \\ \hline
\end{tabular}
\end{center}
\vspace{-0.5cm}
\caption{\small Results of common existing detection and recognition pre-trained methods. Lower ($\downarrow$) results imply a better anonymization. Upper ($\uparrow$) results imply a better detection.}
\label{tab:detection_id}
\end{table}
\footnotetext{In the original version of the paper, we did a calculation error with the original numbers being 99.9 and 98.7.}

%% file: tables/table_lwf.tex
\begin{table}[]
\begin{center}
\begin{tabular}{lll}
\hline
{De-ID method} & {VGGFace2 $(\downarrow)$} & {CASIA $(\downarrow)$}  \\ \hline
Original               &  0.986 $\pm 0.010$     &    0.965 $\pm 0.016$   \\ 
Gafni et al. \cite{DBLP:conf/iccv/Gafni19}          &  0.038 $\pm 0.015$     &    0.035 $\pm 0.011$     \\  
Ours                   &  $ \mathbf{0.034 \pm 0.016} $     &   $ \mathbf{0.019 \pm 0.008}  $   \\ 
\hline

\end{tabular}
\end{center}
\vspace{-0.5cm}
\caption{\small Comparisons with SOTA in LWF dataset. Lower ($\downarrow$) identification rates imply better anonymization.}
\label{tab:lwf}
\end{table}

%% file: supp_arxiv.tex
\twocolumn[{%
\renewcommand\twocolumn[1][]{#1}%
\vspace{-3em}
\begin{center}
    \centering
    \Large{\textbf{--- Supplementary material ---}}
    
\end{center}%
\vspace{1em}
}]

\setcounter{section}{0}

\begin{abstract}
In this supplemental document, we compare the generation quality of our method to different baselines (Section \ref{sec:fid_comp}), qualitatively show the diversity of our generation method (Section \ref{sec:id_diversity}), 
demonstrate results on facial occlusions (Section \ref{sec:no_glasses}), explain the limitations of our model (Section \ref{sec:failure}), and detail the architecture of our models (Section \ref{sec:arch_nets}).
\end{abstract}

\section{FID of baseline methods}\label{sec:fid_comp} 
In Table \ref{tab:quality} we show quantitative results on the quality of the generated images. We use the \textit{FID} score \cite{DBLP:conf/nips/HeuselRUNH17}, a metric that compares the statistics of generated samples to those of real samples. The lower the \textit{FID}, the better. To quantify the quality of generated samples, they are first embedded into a feature space given by (a specific layer) of Inception Net. Then, viewing the embedding layer as a continuous multivariate Gaussian, the mean and covariance are estimated for both the generated data and the real data. The Fréchet distance between these two Gaussians is then used to quantify the quality of the samples:

\begin{equation}
\begin{split}
FID(x, g) = ||\mu_x - \mu_g||_2^2 + Tr(\Sigma_x + \Sigma_g - 2(\Sigma_x \Sigma_g)^{1/2})
\end{split}
\end{equation}

where $(\mu_x, \Sigma_x)$, and $(\mu_g, \Sigma_g)$ are the mean and covariance of the sample embeddings from the data distribution and model distribution, respectfully. The authors show that the score is consistent with human judgment and is robust to noise. Finally, \textit{FID} can detect intra-class mode dropping, \eg a model that generates only one image per class will have a high FID.

Our method reaches a very low \textit{FID} score quantitatively showing that the quality of the generated images is very high. Pixelization methods unsurprisingly reach a very high \textit{FID} score. The higher the number of pixels merged together is, the higher is the \textit{FID} score. Similar behavior is seen for the blurring methods where the higher the level of blur is, the higher is the \textit{FID} score.

We train GAN-based image translation methods \cite{DBLP:conf/cvpr/IsolaZZE17,DBLP:conf/iccv/ZhuPIE17} where the source domain are the landmarks and the target domain are the images. We were surprised to see that the \textit{FID} score of image translation methods is very high, comparable to blurring methods. We investigate this by checking the visual quality of the generated images. We see that the models learn to generate only slight variations of the same face (that can be considered an average face). We conjecture that the main reason for this problem is the sparse signal of the source domain. We show a qualitative evaluation of the baseline methods in Fig. \ref{fig:baseline} where our method is the only one that generates realistically looking images.

\input{tables/table_inception}

\section{Diversity of the generated images}\label{sec:id_diversity} 

In Fig. 1 of the main paper, we showed triplets of images, with the first image in the triplet being the source image, and the other two images being different anonymized versions of it.

In Fig. \ref{fig:id_var} we perform a similar experiment, but this time instead of showing only $2$ anonymized versions of the source image, we present $9$ different versions of it. We see that the generated images have still different identities that are sufficiently different from each other.

\section{Removing glasses and mustache} \label{sec:no_glasses} 

\begin{figure}[t]
\begin{center}
   \includegraphics[width=0.4\textwidth, trim={0.0cm 0.cm 0.cm 0.0cm},clip]{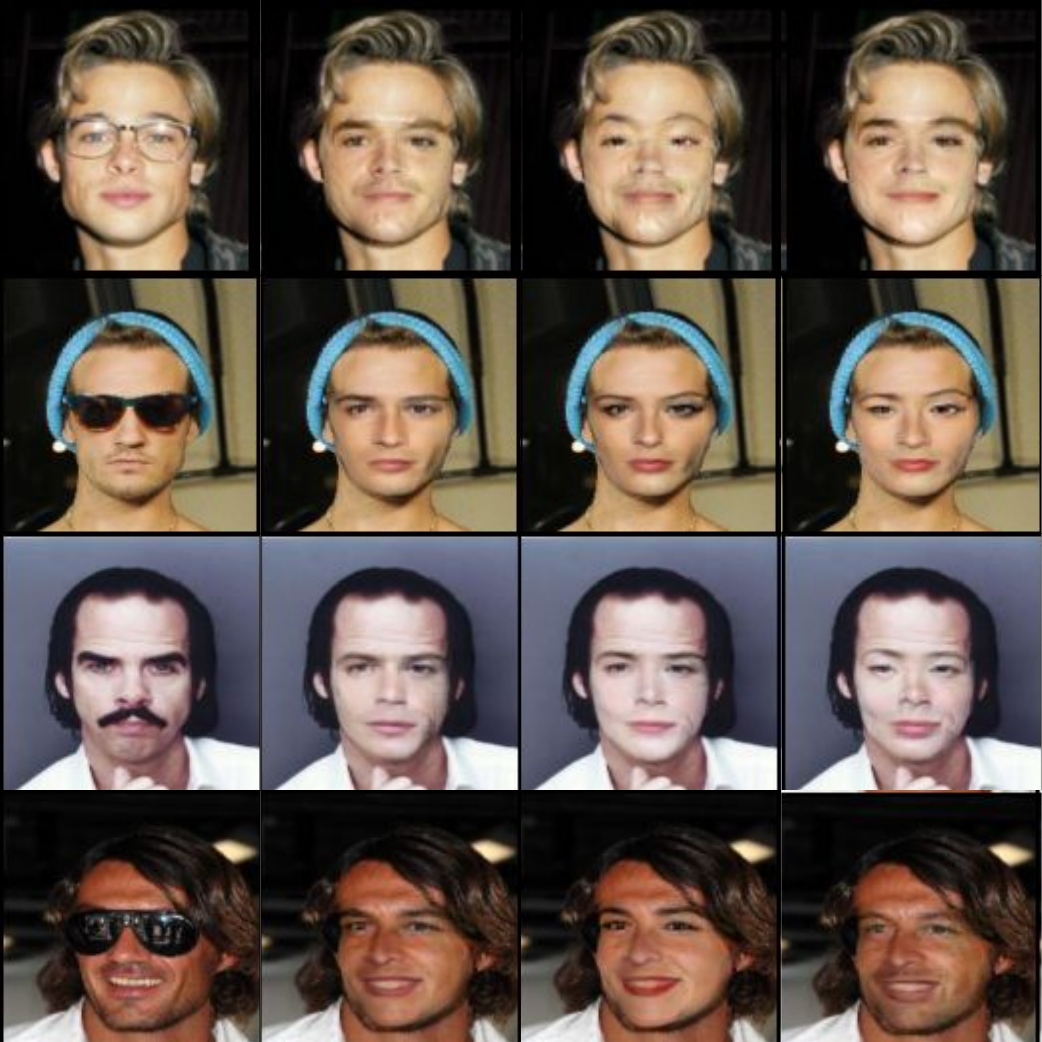}
\end{center}
\vspace{-1.5em}
   \caption{First column contains source images, all other columns are anonymized versions of our method with different control vectors.}
\label{fig:glasses}
\vspace{0.3cm}
\end{figure}

\begin{figure*}[p]
\centering
\includegraphics[width=0.7\textwidth, trim={0cm 0cm 0.cm 0.cm},clip]{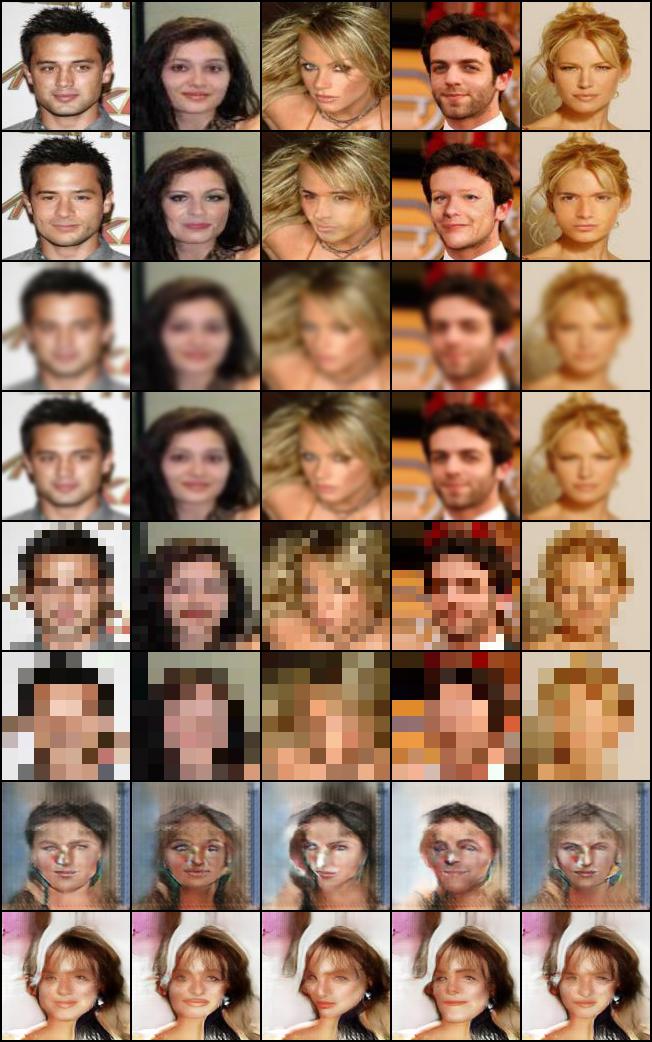}
  \caption{Our method compared with other anonymization baselines. From up to down: Original, CIAGAN, Blur 17 by 17, Blur 9 by 9,  Pixelization 16 by 16, Pixelization 8 by 8, Pix2Pix, CycleGAN.}
  \label{fig:baseline}
\end{figure*}

\begin{figure*}[p]
\begin{center}
   \includegraphics[width=0.95\textwidth, trim={0.0cm 0.cm 0.cm 0.0cm},clip]{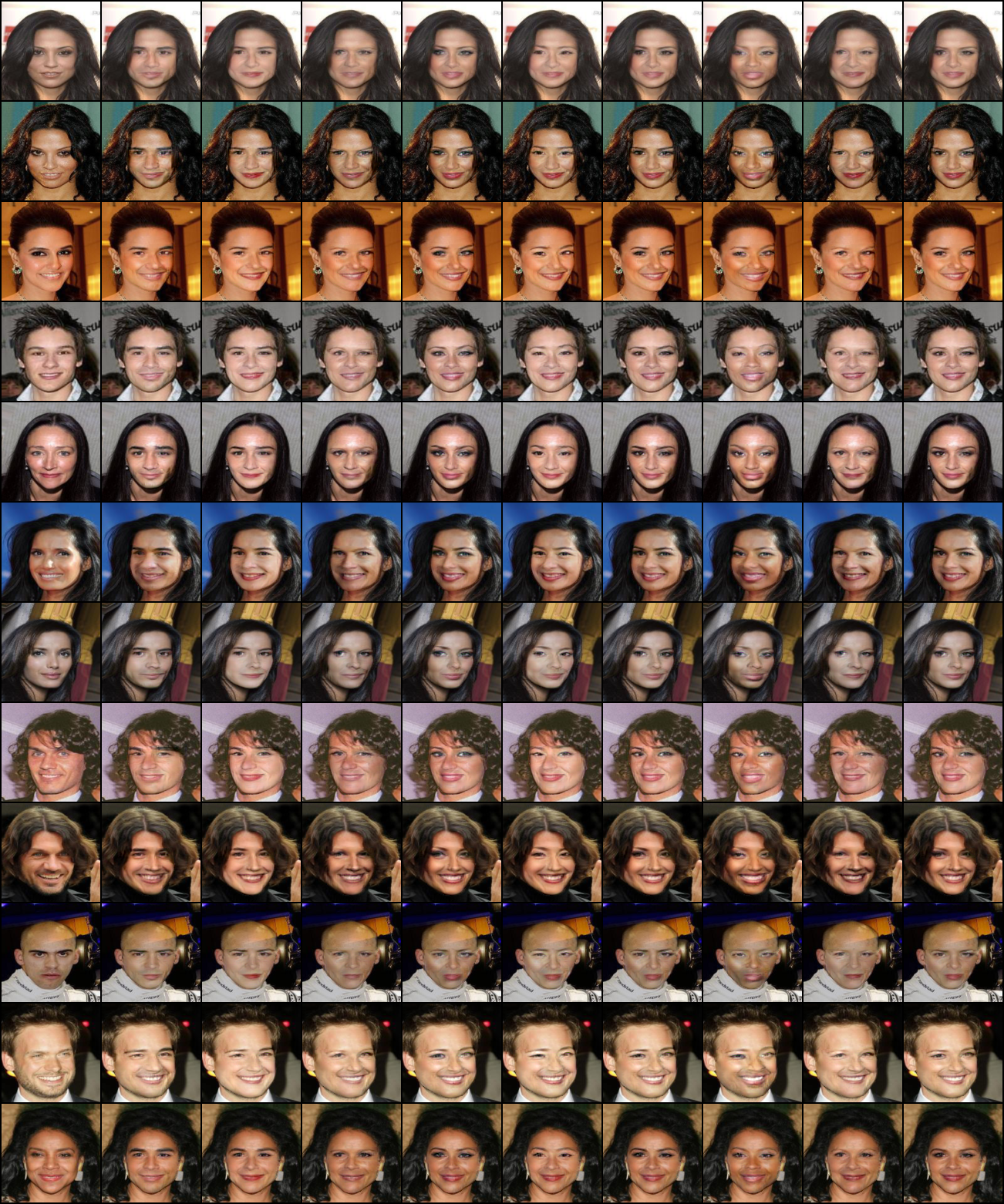}
\end{center}
\vspace{-1.5em}
   \caption{The images in the first column are the source images, all the other images are anonymized versions of the source images, where the anonymization process uses different control vectors.}
\label{fig:id_var}
\end{figure*}

In Fig. \ref{fig:glasses} we perform an experiment where the source images contain faces of people with glasses, or with a mustache. Considering that our generator uses as input the landmarks of the face (instead of the entire face), it has no knowledge whatsoever that the source image might contain glasses. Unsurprisingly, the generated images do not contain glasses or mustache. Note that the quality of the generated images does not suffer and during the blending process, the generator inpaints the region where the source image contained the glasses.

\section{Limitations}\label{sec:failure} 
A weakness of all current de-identification methods \cite{ren2018privacy, DBLP:conf/cvpr/SunMOGSF18, DBLP:conf/iccv/Gafni19} is that they need the original faces to be initially detected before they can be anonymized. Consequently, any face that has not been detected can not be anonymized. Thus, all the aforementioned methods are not deployable in systems where the anonymization must be guaranteed. Our method suffers from a similar issue, if a face does not get detected, then its landmark will not get generated, and so the model will not generate the anonymized version of the face. As future work, we plan on investigating this problem in two directions: deploying an ensemble of detection networks to minimize the probability of faces not getting detected; and anonymizing entire images without the need of detecting the original faces on them. 

Another limitation of landmarks usage is the occurrence of occlusions in front of a face. Since the generated face is based on the landmarks, these occlusions are going to be removed (\eg glasses as in Fig. \ref{fig:glasses}). The problem can be resolved by detecting such occlusions and treating them as part of the background mask.

Finally, like in other deep learning anonymization frameworks, the more different the images are to the images of the training dataset, the worse is the quality of the generation. CelebA dataset offers multiple images per identity  with  good  quality  but with  a  significant  bias towards frontal faces (since they are using photos of celebrities). Consequently, our method performs best when used in similar datasets. For our model to perform as well in extreme poses, it needs to be additionally trained in a dataset containing such poses.

\section{Network architecture} \label{sec:arch_nets} 

In Table \ref{net:generator} we show the architecture of the generator. The generator uses an encoder-decoder architecture, and receives as input a $6$-dimensional image that is created by concatenating the \textit{landmark image} with the \textit{masked background image}. It encodes the input image into a $3$-dimensional tensor that has $256$ channels. In the bottleneck, the encoded image is concatenated with the identity embedding (that is an output of the transposed convolution network). Finally, the network decodes the combined embedding, to produce the anonymized version of the source image.

In Table \ref{net:discr} we show the architecture of the discriminator. The network receives as input a $3$-dimensional image that is created by combining the \textit{generated face} with the \textit{masked background image}. Then it uses a series of residual blocks. Finally it uses $2$ fully-connected layers. The network has the same architecture as the siamese network that we use for identity guidance.

In Table \ref{net:embed} we show the architecture of the embedding network. The network receives as input the label of the desired identity (given in one-hot format) and produces a $3$-dimensional tensor. This tensor is fed into the bottleneck of the generator.

Finally, we give the architecture of the residual blocks used in all the networks. The architectures of the "residual block down", "residual block up" and "residual block" are given on Tables \ref{net:resblock_down}, \ref{net:resblock_up} and \ref{net:resblock} respectively.

\input{tables/net_generator.tex}
\input{tables/net_discr.tex}
\input{tables/net_embed.tex}
\input{tables/net_residual.tex}

%% file: tables/table_inception.tex
\begin{table}[]
\begin{center}
\begin{tabular}{ll}  \hline
Model    & FID ($\downarrow$)  \\  \hline
Pixelization 8 by 8  & 510.83   \\  
Pixelization 16 by 16  & 318.43   \\  
Blur 9 by 9 & 54.50   \\  
Blur 17 by 17   & 152.03   \\  \hline
Pix2Pix \cite{DBLP:conf/cvpr/IsolaZZE17}   & 121.41   \\  
CycleGAN \cite{DBLP:conf/iccv/ZhuPIE17}   & 185.26   \\  \hline
Ours  &  \textbf{2.08}  \\  \hline
\end{tabular}
\end{center}
\caption{\small Quality and diversity measurements of different methods.}
\label{tab:quality}
\vspace{0.5cm}
\end{table}

%% file: tables/net_generator.tex
\begin{table}[h]
\begin{center}
\begin{tabular}{cc}
\hline
\multicolumn{2}{c}{\textbf{Generator}}                                             \\ \hline
\multicolumn{1}{c|}{\textbf{Layers}}                        & \textbf{Output size} \\ \hline
\multicolumn{1}{c|}{Input}                                  & 6 x 128 x128         \\
\multicolumn{1}{c|}{Residual Block Down}                    & 32 x 64 x 64         \\
\multicolumn{1}{c|}{Residual Block Down}                    & 64 x 32 x 32         \\
\multicolumn{1}{c|}{Residual Block Down}                    & 128 x 16 x 16        \\
\multicolumn{1}{c|}{Residual Block Down}                    & 256 x 8 x 8          \\
\multicolumn{1}{c|}{Residual Block Down}                    & 256 x 4 x 4          \\
\multicolumn{1}{c|}{Concatenate with the ID embedding}      & 512 x 4 x 4          \\
\multicolumn{1}{c|}{3x3 stride1 conv + ReLU}                & 256 x 4 x 4          \\
\multicolumn{1}{c|}{Residual Block}                         & 256 x 4 x 4          \\
\multicolumn{1}{c|}{Residual Block}                         & 256 x 4 x 4          \\
\multicolumn{1}{c|}{Residual Block}                         & 256 x 4 x 4          \\
\multicolumn{1}{c|}{Residual Block}                         & 256 x 4 x 4          \\
\multicolumn{1}{c|}{Residual Block Up}                      & 256 x 8 x 8          \\
\multicolumn{1}{c|}{Residual Block Up}                      & 128 x 16 x 16        \\
\multicolumn{1}{c|}{Residual Block Up}                      & 64 x 32 x 32         \\
\multicolumn{1}{c|}{Residual Block Up}                      & 32 x 64 x 64         \\
\multicolumn{1}{c|}{Residual Block Up}                      & 16 x 128 x 128       \\
\multicolumn{1}{c|}{3x3 stride1 conv}                       & 3 x 128 x 128        \\ \hline
\end{tabular}
\vspace{-0.5cm}
\end{center}
\caption{\small The network architecture of our generator.}
\vspace{0.5cm}
\label{net:generator}
\end{table}

%% file: tables/net_discr.tex
\begin{table}[h]
\begin{center}
\begin{tabular}{cc}
\hline
\multicolumn{2}{c}{\textbf{Discriminator}}                      \\ \hline
\multicolumn{1}{c|}{\textbf{Layers}}     & \textbf{Output size} \\ \hline
\multicolumn{1}{c|}{Input}               & 3 x 128 x128         \\
\multicolumn{1}{c|}{Residual Block Down} & 32 x 64 x 64         \\
\multicolumn{1}{c|}{Residual Block Down} & 64 x 32 x 32         \\
\multicolumn{1}{c|}{Residual Block Down} & 128 x 16 x 16        \\
\multicolumn{1}{c|}{Residual Block Down} & 256 x 8 x 8          \\
\multicolumn{1}{c|}{Residual Block Down} & 512 x 4 x 4          \\
\multicolumn{1}{c|}{FC + LeakyReLU}      & 1024             \\
\multicolumn{1}{c|}{FC + LeakyReLU}      & 1                \\ \hline
\end{tabular}
\vspace{-0.5cm}
\end{center}
\caption{\small The network architecture of our discriminator.}
\vspace{0.5cm}
\label{net:discr}
\end{table}

%% file: tables/net_embed.tex
\begin{table}[]
\begin{center}
\begin{tabular}{cc}
\hline
\multicolumn{2}{c}{\textbf{Transposed Convolutional Neural Network}}                                   \\ \hline
\multicolumn{1}{c|}{\textbf{Layers}}                      & \textbf{Output size} \\ \hline
\multicolumn{1}{c|}{Input}                                & N                \\
\multicolumn{1}{c|}{(FC + LeakyReLU) x 7}                 & 512              \\
\multicolumn{1}{c|}{Reshape}                              & 32 x 4 x 4           \\
\multicolumn{1}{c|}{3x3 stride 1 conv + LeakyReLU + IN}   & 64 x 4 x 4           \\
\multicolumn{1}{c|}{3x3 stride 1 conv + LeakyReLU + IN}   & 128 x 4 x 4          \\
\multicolumn{1}{c|}{3x3 stride 1 conv + LeakyReLU + IN}   & 256 x 4 x 4          \\
\multicolumn{1}{c|}{Concatenate with the landmark embedding} & 512 x 4 x 4          \\ \hline
\end{tabular}
\vspace{-0.5cm}
\end{center}
\caption{\small The architecture of our transposed convolutional neural network.}
\vspace{0.5cm}
\label{net:embed}
\end{table}

%% file: tables/net_residual.tex
\begin{table}[]
\begin{center}
\begin{tabular}{cc}
\hline
\multicolumn{2}{c|}{\textbf{Residual Block Down}}                  \\ \hline
\multicolumn{2}{c}{Input}                                          \\
\multicolumn{1}{c|}{3x3 stride1 conv + ReLU} & 1x1 stride1 conv    \\
\multicolumn{1}{c|}{3x3 stride1 conv}        & 2x2 average pooling \\
\multicolumn{1}{c|}{2x2 average pooling}     &                     \\
\multicolumn{2}{c}{Summation}                                      \\
\multicolumn{2}{c}{Instance Normalization}                         \\ \hline
\end{tabular}
\vspace{-0.5cm}
\end{center}
\caption{The network architecture for the residual block down module. }
\vspace{0.5cm}
\label{net:resblock_down}
\end{table}

\begin{table}[]
\begin{center}
\begin{tabular}{cc}
\hline
\multicolumn{2}{c|}{\textbf{Residual Block Up}}                                          \\ \hline
\multicolumn{2}{c}{Input}                                                                \\
\multicolumn{1}{c|}{IN + ReLU}                    & \multicolumn{1}{l}{2x2 Upsample}     \\
\multicolumn{1}{c|}{2x2 Upsample}                 & \multicolumn{1}{l}{1x1 stride1 conv} \\
\multicolumn{1}{c|}{3x3 stride1 conv + IN + ReLU} &                                      \\
\multicolumn{1}{c|}{3x3 stride1 conv}             &                                      \\
\multicolumn{1}{c|}{2x2 average pooling}          &                                      \\
\multicolumn{2}{c}{Summation}                                                            \\
\multicolumn{2}{c}{Instance Normalization}                                                                   \\ \hline
\end{tabular}
\vspace{-0.5cm}
\end{center}
\caption{The network architecture for the residual block up module.}
\vspace{0.5cm}
\label{net:resblock_up}
\end{table}

\begin{table}[]
\begin{center}
\begin{tabular}{c}
\hline
\multicolumn{1}{c|}{\textbf{Residual Block}} \\ \hline
Input                                        \\
3x3 stride1 conv + IN + ReLU                 \\
3x3 stride1 conv + IN                        \\
Summation with Input                         \\ \hline
\end{tabular}
\vspace{-0.5cm}
\end{center}
\caption{\small The network architecture for the residual block module. IN stands for Instance Normalization \cite{DBLP:journals/corr/UlyanovVL16}}
\vspace{0.5cm}
\label{net:resblock}
\end{table}